\pdfoutput=1
\documentclass[11pt]{article}

\usepackage{acl}

\usepackage{times}
\usepackage{latexsym}
\usepackage{multirow}
\usepackage{xspace}
\usepackage{CJKutf8}
\usepackage{hyperref}
\usepackage[T1]{fontenc}
\usepackage[utf8]{inputenc}

\usepackage{microtype}
\usepackage{graphicx} 
\usepackage{booktabs}

\newcommand\BERTbase{BERT$_{\tiny\textsc{BASE}}$\xspace}
\newcommand\Rbase{RoBERTa$_{\tiny\textsc{BASE}}$\xspace}
\newcommand\Rlarge{RoBERTa$_{\tiny\textsc{LARGE}}$\xspace}

\title{Information Extraction and Human-Robot Dialogue towards Real-life Tasks: A Baseline Study with the MobileCS Dataset}

\author{Hong Liu$^{1,3,\dagger}$, Hao Peng$^{1,3,\dagger}$, Zhijian Ou$^{1,3}$\thanks{~~Corresponding author, $^{\dagger}$Equal contribution}, Juanzi Li$^{1,3}$, Yi Huang$^{2,3}$, Junlan Feng$^{2,3}$ \\
  $^{1}$Tsinghua University, Beijing, China \\
  $^{2}$China Mobile Research Institute, Beijing, China \\
  $^{3}$Tsinghua University-China Mobile Communications Group Co., Ltd. Joint Institute, Beijing, China \\
  \texttt{\{liuhong21,peng-h21\}@mails.tsinghua.edu.cn},\\
  \texttt{\{ozj,lijuanzi\}@tsinghua.edu.cn},\\
  \texttt{\{huangyi,fengjunlan\}@chinamobile.com}
}

\begin{document}
\maketitle
\begin{abstract}
% Previous task-oriented dialogue datasets are collected from the conversations of crowd workers, where specific instructions are provided for workers in each conversation. 
Recently, there have merged a class of task-oriented dialogue (TOD) datasets collected through Wizard-of-Oz simulated games.
However, the Wizard-of-Oz data are in fact simulated data and thus are fundamentally different from real-life conversations, which are more noisy and casual.
Recently, the SereTOD challenge is organized and releases the MobileCS dataset, which consists of real-world dialog transcripts between real users and customer-service staffs from China Mobile. Based on the MobileCS dataset, the SereTOD challenge has two tasks, not only evaluating the construction of the dialogue system itself, but also examining information extraction from dialog transcripts, which is crucial for building the knowledge base for TOD.
This paper mainly presents a baseline study of the two tasks with the MobileCS dataset. We introduce how the two baselines are constructed, the problems encountered, and the results.
We anticipate that the baselines can facilitate exciting future research to build human-robot dialogue systems for real-life tasks.
% In this Wizard-of-Oz data-collection method leads to a gap between collected data and dialogues in real life. To address this issue, a new corpora, MobileCS, coming from real-world dialogue transcripts between real users and customer-service staffs from China Mobile was proposed. Compared to previous datasets, MobileCS is much more challenging with spoken noise and redundant turns in real-life conversations. To establish baseline systems on such dataset is non-trivial and deserves detailed descriptions.
\end{abstract}

\section{Introduction}
\label{sec:intro}
Building human-robot dialogue systems is an important research question not only for artificial intelligence applications but also for artificial intelligence itself.
In the Turing test, if the human evaluator finds that human-robot dialogue and human-human dialogue are indistinguishable, the robot would be said to exhibit intelligent behaviour and pass the test \cite{Turing1950}.
So presumably, the best strategy to build an intelligent dialogue system may be to train the system over a large
amount of real human-to-human conversations to mimic human behaviors.
This approach was once pursued and several human-human dialogue datasets have been released, such as the Twitter dataset \cite{Ritter2010UnsupervisedMO}, the Reddit
conversations \cite{Schrading2015AnAO}, and the
Ubuntu technical support corpus \cite{Lowe2015TheUD}.
It is argued in \cite{budzianowski-etal-2018-multiwoz} that the lack of grounding conversations onto
an existing knowledge base (KB) limits the
usability of the systems developed over these human-human dialogue datasets.

So a class of Wizard-of-Oz simulated games have emerged to collect human-human conversations \cite{wen-etal-2017-network, el2017frames, budzianowski-etal-2018-multiwoz, zhu2020crosswoz, quan-etal-2020-risawoz}, particularly for task-oriented dialogue (TOD) systems which help users accomplish specific goals such as finding restaurants or booking flights and usually require a task-related KB.
In the Wizard-of-Oz set-up, through random sampling based on an ontology and a KB (both are pre-defined), a task template is created for each dialogue session between two crowd workers. One worker acts as the role of a user and the other performs the role of a clerk (i.e. the system side).
In practice, multiple workers may contribute to one dialogue session.
In this way, annotations of belief states and systems acts become easy, and grounding conversations onto the KB is realized.

% Recently, there are increasing interests in building task-oriented dialogue (TOD) systems which help users accomplish specific goals, such as finding restaurants or booking flights. Therefore, large annotated TOD datasets are always in demanding to facilitate the development of TOD systems.

% Previous large TOD datasets, whether composed of written conversations or spoken  conversations, are either collected through WOZ \cite{wen-etal-2017-network, el2017frames, budzianowski-etal-2018-multiwoz, zhu2020crosswoz, quan-etal-2020-risawoz}, or collected by converting machine-generated outlines to natural languages using crowd workers \cite{shah2018building, rastogi2020towards, lee2022sgd}. 

However, dialogue data collected in the Wizard-of-Oz set-up are in fact simulated data and thus are fundamentally different from real-life conversations.
During the Wizard-of-Oz collection, specific instructions (e.g., goal descriptions for the user side and task descriptions for the system side) are provided for crowd workers to follow.
% This is different from the real-life scenarios.
In contrast, real-life dialogues are more casual and free-style, without instructions. Even with some goals in mind, chit-chat or redundant turns are often exist in real-life conversations, e.g., asking for repeating or confirming key information. In some sense, we could say that real-life dialogues are more \emph{noisy}.
Moreover, spoken conversations in real-world have a distinct style with those well-written conversations and are full of extra noise from grammatical errors, influences or barge-ins \cite{9688274}. 
For building dialogue systems that are more applicable to real-life tasks, real human-human dialogue datasets with grounding annotations to KBs are highly desirable. 

Recently, the SereTOD challenge is organized \cite{ou2022achallenge} and releases a new human-human dialogue dataset, called the MobileCS (Mobile Customer Service) dataset. It consists of real-world dialog transcripts between real users and customer-service staffs from China Mobile.
Based on the observation and analysis of those dialogue transcripts, a schema is summarized to our best\footnote{How to build an ``optimal'' schema for a real-life task is still an open research problem. Further investigation of the schema for the MobileCS dataset is an interesting future work.}, according to which about 10,000 dialogues are annotated with entities, attribute triples and speaker's intents for every turn. 
The annotated part of the MobileCS dataset is randomly split into a train, development and test set, which consists of 8975, 1025 and 962 dialogues, respectively.

Based on the MobileCS dataset, the SereTOD challenge not only evaluates the construction of the dialogue system itself (Task 2), but also examines information extraction from dialog transcripts (Task 1), which is crucial for building the KB.
The MobileCS data are more noisy and challenging, as compared to previous Wizard-of-Oz data. It is non-trivial to establish baseline systems on such dataset.
% This paper will introduce the baseline systems of the two tasks, the problems encountered in the construction process of baseline systems and the corresponding solutions.
This paper mainly presents a baseline study of the two tasks with the MobileCS dataset.
Two baseline systems are constructed for the two tasks respectively, which both are released as open source\footnote{https://github.com/SereTOD/SereTOD2022} and provided to the participating teams in the SereTOD challenge.
We introduce how the two baselines are constructed, the problems encountered, and the results.
The results clearly show the challenge for information extraction and human-robot dialogue, when trained and tested on real human-human data.
We anticipate that the baselines can facilitate exciting future research to build human-robot dialogue systems for real-life tasks.

% MobileCS (Mobile Customer-Service) dataset is such a dataset,  which comes from real-world dialogue transcripts between real users and customer-service staffs from China Mobile. Based on the observation and analysis of those dialogue transcripts, a schema was summarized, according to which dialogues are annotated with entities, attribute triples and speaker's intents for every turn.

% Two related tasks, information extraction from dialog transcripts and task-oriented dialog system, are proposed to promote dialogue research in real life. Compared to previous TOD datasets, MobileCS is much more challenging with spoken noise and redundant turns in real-life conversations. To establish baseline systems on such dataset is non-trivial, with many problems never studied before. This paper will introduce the baseline systems of the two tasks, the problems encountered in the construction process of baseline systems and the corresponding solutions.

\section{Related Work}
\subsection{Dialogue Datasets}
According to \citet{budzianowski-etal-2018-multiwoz}, existing dialog datasets (whether task-oriented or not) can be grouped into three categories: machine-to-machine, human-to-machine, and human-to-human. The machine-to-machine datasets may ensure full coverage of all possible dialogue outcomes within a certain domain, but they do not consider noisy conditions in real life, which poses a risk of a mismatch between training data and real interactions. The human-to-machine datasets, however, depend on the provision of an existing working dialogue system, which limits the practicality of the datasets. 
The human-to-human datasets address the problems in the above two classes of datasets. However, previous human-to-human datasets lack knowledge base and explicit goal in the conversation, making that systems trained with these corpus struggle in generating consistent
and diverse responses \cite{li2016diversity}.

It is non-trivial to collect a TOD dataset with knowledge base and user goals. Previous TOD datasets are either collected through Wizard-of-Oz simulated games \cite{wen-etal-2017-network, el2017frames, budzianowski-etal-2018-multiwoz, zhu2020crosswoz, quan-etal-2020-risawoz}, or collected by converting machine-generated outlines to natural languages using crowd workers \cite{shah2018building, rastogi2020towards, lee2022sgd}. However, during the collection of these previous datasets, specific instructions are provided for crowd workers, which is different from real-life conversation scenarios and leads to a gap between collected data and dialogues in real-life. The MobileCS dataset, introduced in SereTOD Challenge, comes from real-world dialogue transcripts and represents a step advancing to remedy the above deficiencies.

\subsection{Dialogue Information Extraction}
Dialogue information extraction is the task of extracting structured information, e.g., entities and attributes, from dialogue transcripts. 
Different from the traditional information extraction in general domain text~\citep{sarawagi2008information, li2020survey, han2020more}, dialogue transcirpts are more verbalized and loose with more irregular expressions and grammar errors. Previous works have explored how to extract user information~\citep{catizone2010using, hirano2015user, wu2019getting}, clinical information~\citep{kannan2018semi, peng2021dialogue}, and relations between speakers and mentioned entities in dialogues~\citep{yu2020dialogue, jia2021ddrel}. 
However, there is no previous work focusing on extracting information on real-world dialogue transcripts between real users and customer-service staffs. 
In the paper, we develop a 
modern dialogue information extraction baseline, based on the MobileCS dataset, which contains dialogue transcripts from China Mobile.

\subsection{Task-oriented Dialogue System}
The methodology for building TOD systems is gradually advancing from separate training of individual modules \cite{williams2016dialog,mrkvsic2017neural, dai2018tracking} to the end-to-end (E2E) trainable approach \cite{wen2017a, liu2017end, lei2018sequicity, fsdm, zhang2020task,gao2020paraphrase,zhang-etal-2020-probabilistic}.
In early E2E methods, the sequential turns of a dialog are modeled with LSTM-based backbones.
Recently, self-attention based Transformer neural networks \cite{vaswani2017attention} have shown their superiority in capturing long-term dependencies over LSTM based networks. Transformer based pretrained language models (PLM), such as GPT2 \cite{radford2019gpt2} and T5 \cite{JMLR:v21:20-074}, have been leveraged to build generative E2E TOD systems in the pretraining-and-finetuning paradigm, which have shown improved performance over LSTM-based ones. Examples include GPT2-based SimpleTOD \cite{hosseini2020simple}, SOLOIST \cite{peng2020etal}, AuGPT \cite{kulhanek2021augpt} and UBAR \cite{yang2021ubar}, and T5-based PPTOD \cite{su2021multitask} and MTTOD \cite{lee-2021-improving-end}, among others.
However, these previous TOD systems are mainly examined on simulated data collected by crowd workers.
It is not clear what the potential performance of the current methodology of building TOD systems is in real-life tasks.
In this paper, we present our effort to answer this question, by developing a TOD system on the MobileCS data, which are from real-life customer-service. 
% limits their practicality. In this paper, we try to build a TOD system on the dialogues from real life, which is distinct from previous systems.

% However, dialogues are abundant resource and usually contain valuable information. 

\section{MobileCS Dialogue Dataset}
\label{sec:data}
The MobileCS dialogue dataset contains 10,000 dialogue labeled by crowd-sourcing and around 90,000 unlabeled dialogues. For each dialogue turn, the annotations consist of entities, attribute triples, and speaker's intents within the scope of the schema. Another around 1000 dialogs are put aside as the test data.
More detailed information about the MobileCS dataset can be found in the challenge description paper for the SereTOD challenge \cite{ou2022achallenge}. 
% \footnote{http://seretod.org/SereTOD\_Challenge\_Description\_v2.0.pdf}. 

The two tasks defined over the MobileCS dataset for the SereTOD challenge require different annotations.
For information extraction (Task 1), the annotations of entities and attribute triples are needed for training and evaluating the system. For TOD system construction (Task 2), user intents, system intents and a local knowledge base (\emph{local KB}, which covers personal information and relevant public knowledge in a dialogue) are required. 
A \emph{global KB}, which covers and fuses all public knowledge and all personal information in the MobileCS domain, is difficult to obtain during the research phase.
Thus, the SereTOD challenge introduces the concept of a local KB, which could be viewed as being composed of the relevant local snapshots from the global KB for each dialog.
The local KB is obtained automatically by integrating all the annotations of entities and attributes into a sequence of entities\footnote{An interesting future problem is to study the quality of the local KBs constructed in such a way and their influence on the performance of the dialogue system.}. Besides, user goals are needed for evaluating the performance of TOD systems in Task 2. Similarly, user goals are obtained automatically by integrating user intents and all the entities and attributes mentioned by the user. The examples of local KB and user goal can be found in Listing
1 in the challenge description paper \cite{ou2022achallenge}.
\paragraph{Data Quality}
The MobileCS data were annotated by two professional data labeling teams according to well documented guidelines as described in \cite{ou2022achallenge}. 
Quality control was enforced by sampling the annotated data and performing crossing check of the annotations by the two teams.
Nevertheless, there still exist annotation errors in such a large dataset. 
Some annotation errors can be corrected by rules. A typical example of errors is the granularity error of entity types. In the schema, entity types have inheritance relationships, for example, ``main package'' inherits from ``package'' and contains all its properties. Therefore, there are quite a few annotation confusions between parent types and child types in the data. To correct those type errors, the most fine-grained type for each entity was selected according to the attributes held by the entity. By combining the schema with manual rules, more annotation errors can be corrected.
The updated MobileCS dataset is called v1.1, which is released in the SereTOD challenge and used in the experiments in this paper.

\section{Tasks}
\subsection{Task 1: Information Extraction from Dialog Transcripts}
% Task-oriented dialogue systems require 
% natural language understanding and background knowledge to make responses. 
Task 1 aims to extract structured information from real-world dialogue transcripts for constructing KB for the TOD system.
This task consists of two sub-tasks: entity extraction and slot filling. The entity extraction task aims to extract entities, involving
named entity recognition and entity coreference resolution. And the slot filling task aims to extract the attributes and values of entities, and the status of user accounts.
Compared to the information extraction tasks on general domain texts, this task poses more challenges.
First, dialogue transcripts are more verbalized, loose and noisy, which requires more robust models. Second, dialogue transcripts contain more pronouns and referents, some of which even span several rounds. This requires coreference resolution and long context modeling. 

\subsection{Task 2: Task-Oriented Dialog Systems}
The basic task for the TOD system is, for each dialog turn, given the dialog history, the user utterance and the local KB, to predict the user intent, query the local KB and generate appropriate system intent and response according to the queried information. Compared with previous work, this task has the following characteristics.
First, there is no global KB but only a local KB for each dialog, containing all the information in entity and attribute annotations and representing the unique information for each user, e.g., the user's package plan and remaining phone charges.
Second, the user's constraints on entities are relatively simple, e.g., 38M data package, so the customer service system usually can 
confirm the entities that the user refers to in one dialogue turn, without the need of dialogue state accumulation.

\begin{CJK*}{UTF8}{gbsn}
\begin{figure*}[t]
    \centering
    \includegraphics[width=0.98\linewidth]{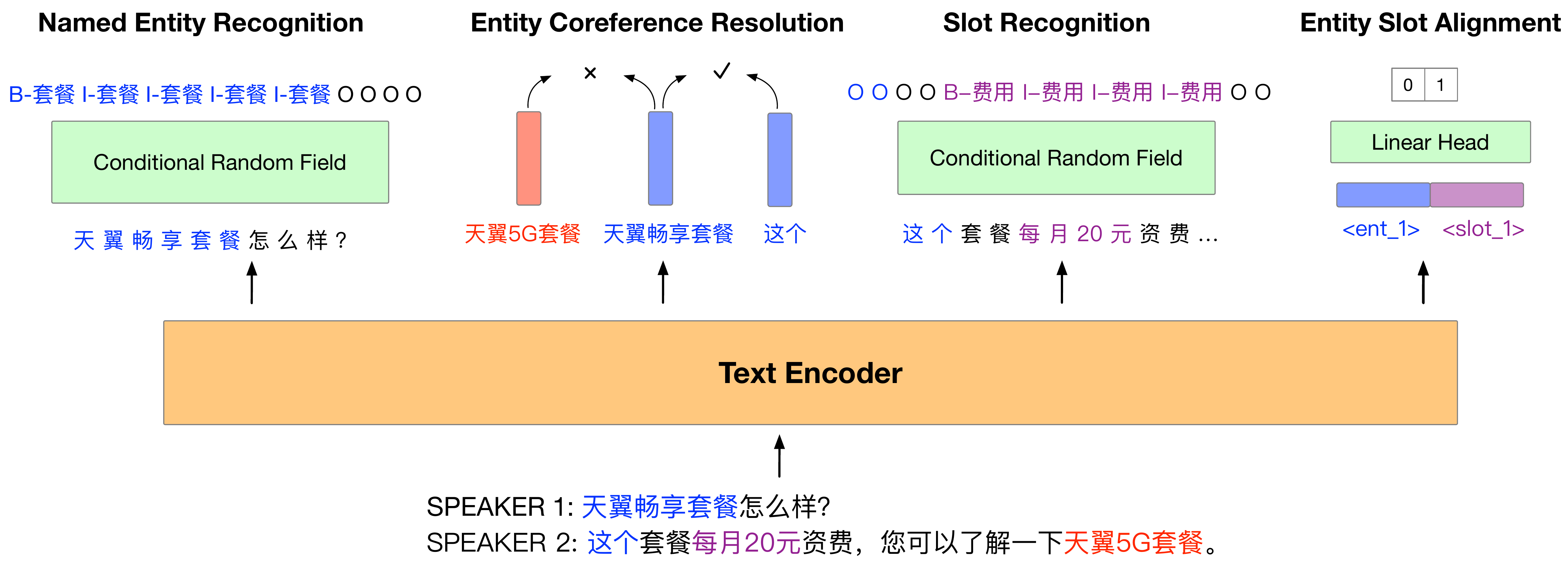}
    \caption{The overall model architecture of the pipeline model for Task 1. For the sub-task entity slot alignment, we utilize marker (e.g., \texttt{<ent\_1>}\textit{entity mention}\texttt{<$\backslash$ent\_1>}) to highlight entities and slots in the original text input. }
    \label{fig:track1_model}
\end{figure*}
\end{CJK*}

\section{Baseline Models}
\subsection{Task 1 Baseline}
Task 1 involves two challenging sub-tasks: entity extraction and slot filling.
Therefore, we design a pipeline method to extract information from dialogue transcripts. For entity extraction, the pipeline is two-step: named entity recognition and entity coreference resolution. For slot filling, the pipeline is also two-step: slot recognition and entity slot alignment. 
For each step, we first utilize a text encoder backbone to encode utterances and then a task-specific module to extract specific information based on the encoded representations of the utterances. In our experiments, we adopt three text encoders: LSTM~\citep{lai2015recurrent}, BERT~\citep{kenton2019bert}, and RoBERTa~\citep{liu2019roberta}.
The overall model architecture is illustrated in Figure~\ref{fig:track1_model}. The hyper-parameters are shown in Table~\ref{tab:hyper-param}.
The details of each step are as follows.

\paragraph{Named Entity Recognition}
First, we utilize a sequence labeling method to extract entity mentions in dialogue transcripts as in ~\citet{yamada2020luke}. Specifically, after encoding utterances, we adopt conditional random field~\citep{crf} on the top of hidden representations to label entity mentions from each utterance of the dialogue transcripts.

\paragraph{Entity Coreference Resolution} After extracting entity mentions from dialogue transcripts, we utilize an entity coreference resolution method to group the mentions that refer to the same entity, as the local KB organizes knowledge in entity level instead of mention level. Specifically, after encoding dialogues, we adopt the dot product between the representation vectors of the two entity mentions as the metric to assess whether two mentions refer to the same entity. The representation vector of an entity mention is defined by the mean pooling of the representations of the tokens of the entity mention, as did in~\citet{yao2019docred}.
We then utilize the binary cross entropy loss as the objective to fine-tune the backbone encoders. 

\paragraph{Slot Recognition}
Slot recognition aims to recognize slots from plain texts, regardless of which entity the slot belongs to. We utilize a sequence labeling method to recognize the slots, i.e., to label certain spans in the utterance as slots, which are the attributes of entities and the status of users.
Specifically, we utilize the same model architecture as in Named Entity Recognition to label slots from each utterance of the dialogue transcripts.

\paragraph{Entity Slot Alignment}
To construct a local KB, the final procedure is to link slots to the corresponding entities. We formulate the task as a sequence classification task. Specifically, we highlight an entity and a slot using special markers and then encode the text to the contextual representation, which is inspired by~\citet{soares2019matching}. We adopt a linear classification head to classify whether the slot corresponds to the entity. 

\begin{table}[]
    \centering
    \small
    \begin{tabular}{l|cc}
    \toprule
    Hyper-parameter & LSTM & PLMs \\
    \midrule
    Learning Rate & $1\times10^{-3}$ & $3\times10^{-5}$ \\ 
    Batch Size & 64 & 64 \\
    Epoch & 20 & 5 \\ 
    \bottomrule
    \end{tabular}
    \caption{Hyper-parameters of fine-tuning LSTM and PLMs (BERT, RoBERTa) on Track1 task.}
    \label{tab:hyper-param}
\end{table}

\subsection{Task 2 Baseline}
\label{sec:task2-baseline}
\paragraph{KB Query}
We need to design a KB query function to help the TOD system access information from the local KB. After observing the dataset, we find that user queries can be divided into three different types. We encapsulate all query scenarios into one function and list their inputs (i.e. the arguments of the query function) and outputs as follows.
\begin{itemize}
    \item Query the attribute of a specified entity. The input is the entity name and the attribute to be queried, the output is the attribute value in the local KB.
    \item Query entities of specified types. The input is entity type, the output is the entity names of this type.
    \item Query the attribute for users. The input is the attribute to be queried, the output is the queried attribute value in the local user profile (part of the local KB).
\end{itemize}
With the above query function, the TOD system can use the predicted user intent to access information from the local KB.

\paragraph{Baseline Architecture} 
%The overview of Task 2 baseline is shown in Figure~\ref{fig:track2-baseline}. 
\begin{figure*}[t]
\centering
	\includegraphics[width=0.95\linewidth]{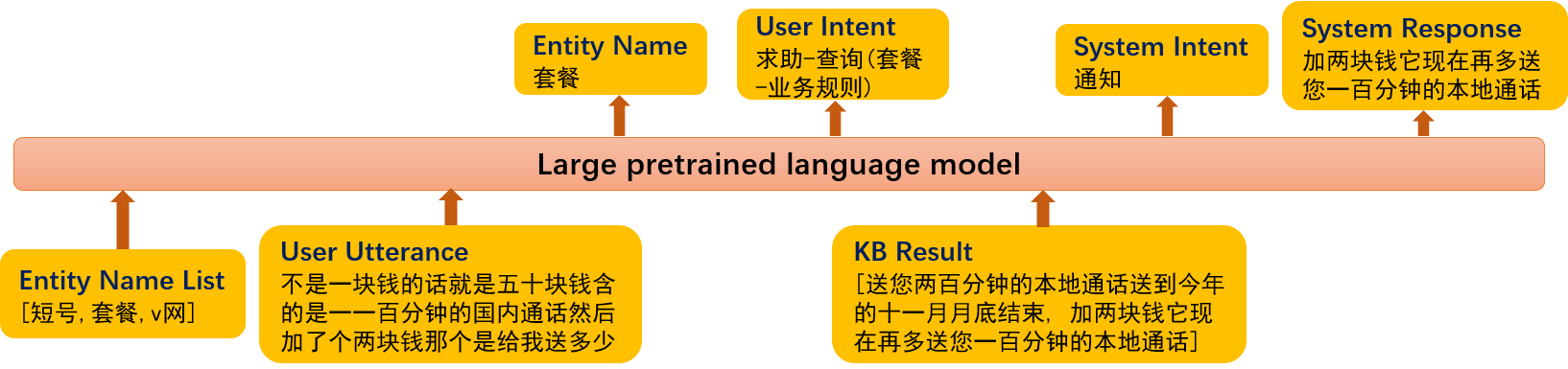}
	\caption{The baseline model architecture for Task 2. Examples are provided under the title of each box.}
	\label{fig:track2-baseline}
\end{figure*}

We divide the TOD system into several sub-tasks. For every dialog turn, the system needs to perform the following steps in order: 1) predict the entity name mentioned or referred to by the user; 2) predict the user intent (including the arguments of the query function); 3) query the local KB using the predicted user intent and obtain the KB result; 4) predict the system intent; 5) predict the system response. Note that there are many pronouns and co-references of entity names, so that the system may not be able to predict correct entity name with only the user utterance in current turn. To solve this problem, dialogue history information is needed. However, in real-life dialogues, the dialogue history is particularly long and contains plenty of characters, which will seriously hurt the training efficiency of the model \cite{Liu2022RevisitingMG}.
Therefore, we maintain a list of entity names mentioned by the user in all previous turns (entity name history) to replace the dialogue history. 
The entity name history and user utterance are fed into the model as the conditioning input to complete the above sub-tasks. 
Similar to \citet{hosseini2020simple, peng2020etal, kulhanek2021augpt,yang2021ubar, Liu2022RevisitingMG}, we employ a sequence generation architecture based on Chinese GPT-2 \cite{GPT2-Chinese} to implement the dialog system, which is depicted in Figure~\ref{fig:track2-baseline}. 
 
\paragraph{Data Analysis} 
As described in Section \ref{sec:intro}, there are chit-chat or redundant turns in real-life dialogues. As observed from MobileCS, we find that these redundant turns can be divided into three cases:
1) one speaker asks for repetition and the other repeats what he/she said before;
2) one speaker confirms information and the other responds to it passively;
3) the user interrupts the agent with some simple interjections, and then the agent continues to speak. Three examples corresponding to the three cases are shown in Table~\ref{tab:cleaning-case}.

These redundant turns are interesting new phenomena revealed from the MobileCS data, which are transcribed from spoken conversations.
Remarkably, the repetition and confirmation may be caused by that the staff did not hear clearly due to the accent or low quality of the user speech.
The interjection is a special feature for spoken dialogues.
However, after transcription of speech, the speech modality is missing, since only text is remained. Thus, the system in textual dialogues receives no relevant input from the user and is thus unable to respond properly. 
%there is no signal to the system indicating when the repetition or confirmation is required, and the system is unable to determine its response when the user only utters one interjection.
We leave further study of this problem for future work. In this work, we perform some pre-processing on the data to reduce the noise brought by the three cases. Specifically, for the first two cases, we simply delete the whole redundant turn (including utterances on both user and system sides) in the dialogue. For the third case, we merge the redundant turn with its previous turn by deleting the user utterance and merge the agent response with the previous one. Finally, we obtain a cleaned dataset with 15\% fewer turns than the original one.

\section{Evaluation}
\subsection{Task 1 Results}
\paragraph{Metrics}
The evaluation metrics are two-fold. The metric for entity extraction is the span-level F1, following previous named entity recognition work~\citep{luke}. The metric for slot filling is the triple-level F1: a predicted \textit{entity-slot-value} triple is correct if and only if the entity, slot and value are all correct. The evaluation for slot filling is a combinatorial optimization problem, as the entity is also predicted. We hence utilize the Hungarian algorithm~\citep{kuhn1955hungarian}
to find a best entity matching between predictions and golden labels before calculating the metric for slot filling.
\paragraph{Results}
The models are trained on the training set for a certain number of epochs (as shown in Table~\ref{tab:hyper-param}), selected according to performance over the dev set, and evaluated on the official test set\footnote{Notably, the challenge leaderboad for Track 1 are ranked by the results tested over 500 dialogues, which is only a subset of the official test set and were held out by the Challenge Organizers and not sent to the teams.}.
The results are shown in Table~\ref{tab:task1_result}. It can be seen that even with powerful pre-trained language models as text encoders, the performance of the baseline model is poorer on the MobileCS dataset, especially for the named entity recognition and slot recognition sub-tasks, as compared to results on other datasets reported in the literature \cite{yamada2020luke}.
These results demonstrate how demanding the MobileCS dataset is, and indicate that extracting structured information from long and loose texts, e.g. dialogue transcripts, remains challenging for existing models, which urges more powerful and robust models.

\begin{CJK*}{UTF8}{gbsn}
\begin{table}[t]
\centering
\resizebox{\linewidth}{!}{
\begin{tabular}{ll}
\toprule
user &成那改成那最便宜那是打打那个长途是多少钱呢\\
system &\textcolor{blue}{呃呃您再说一下}\\
user &\textcolor{blue}{我说改成那种你说那个便宜的是打打那个长途是多少钱一分钟呢}\\
\midrule
user &沈那中学里面\\
system &\textcolor{blue}{沈那中学是吗}\\
user &\textcolor{blue}{对}\\
\midrule
system &三十八我看这边是流量送您六百兆通话送您两百分钟\\
user &\textcolor{blue}{嗯}\\
system &前三个月每个月还送您二十块钱话费和一g的流量\\
\bottomrule
\end{tabular}
}
% \vspace{-0.8em}
\caption{Examples of three types of redundant turns in MobileCS. The redundant utterances are marked in blue.}
\label{tab:cleaning-case}
\end{table}
\end{CJK*}
% \lipsum[1-15]

\begin{table*}[]
    \centering
    \small
    \begin{tabular}{l|r|rrr|c} 
    \toprule
    \multirow{2}{*}{Backbone} & \multirow{2}{*}{F1 (NER)} &\multicolumn{3}{c|}{Golden Labels}  & Pipeline \\
    \cmidrule{3-6}
    &  & $B^3$ (ECR) & F1 (SR) & Acc. (ESA) & F1 (SF) \\
    \midrule
    LSTM~\citep{lai2015recurrent} & $35.02$ & $85.84$ & $43.89$ & $76.84$ & $31.37$ \\ 
    \BERTbase~\citep{kenton2019bert} & $34.21$ & $88.09$ & $46.46$ & $76.50$ & $33.24$ \\
    \Rbase~\citep{liu2019roberta} & $33.74$ & $88.02$ & $45.59$ & $77.32$ & $33.28$ \\
    \Rlarge~\citep{liu2019roberta} & $\mathbf{35.06}$  & $\mathbf{89.42}$ & $\mathbf{46.89}$ & $\mathbf{78.07}$  & $\mathbf{34.95}$ \\
    \bottomrule
    \end{tabular}
    \caption{Experimental results of Task 1 on the official test set, with different text encoder backbones. ``Golden Labels'' means using golden prerequisite labels (e.g. golden entities for entity coreference resolution) for each pipeline step. ``Pipeline'' represents using previous predictions for each pipeline step. The evaluation metric is micro F1 for named entity recognition and entity slot alignment, B-cubed metric~\citep{bagga1998entity} for entity coreference resolution, and accuracy for entity slot alignment.
    NER: Named Entity Recognition. ECR: Entity Coreference Resolution. SR: Slot Recognition. ESA: Entity Slot Alignment. SF: Slot Filling. }
    \label{tab:task1_result}
\end{table*}

\subsection{Task 2 Results}
\paragraph{Metrics} 
In order to measure the performance of TOD systems, both automatic evaluation and human evaluation are conducted. 
For automatic evaluation, metrics include \textbf{Precision/Recall/F1 score, Success rate and BLEU score}.  P/R/F1 are calculated for both predicted user intents and system intents.
Success rate is the percentage of generated dialogs that achieve user goals. Specifically, for each dialogue, we extract the information requested in the user goal from the local KB, then we regard this dialogue as a success if the generated responses contain all the requested information.
BLEU score evaluates the fluency of generated responses by comparing them with oracle responses. %A combined score is computed as BLEU+Success.
%The combined score in is calculated as User intent F1 + System intent F1 + Success + BLEU/50.
For human evaluation, 5 testers (staffs from China Mobile) interacted with the system, and each tester should interact for at least 10 dialogues with the system. The tester would score the system on a 5-point scale (1 to 5) by the following 3 metrics. 
Success measures if the system successfully completes the user goal by interacting with the user.
Coherency measures whether the system’s response is logically coherent with the dialogue context.
Fluency measures the fluency of the system’s response.
\paragraph{Results}
Based on the analysis in Section~\ref{sec:task2-baseline}, we conduct experiments on the original dataset and the cleaned dataset, respectively. The models are trained on the official training set for 40 epochs, and tested on the official dev set. The results are shown in Table~\ref{tab:task2-result}.
\begin{table}[t]
\centering
\resizebox{\linewidth}{!}{
\begin{tabular}{lcccc}
\toprule
Dataset & U-P/R/F1  & S-P/R/F1 &BLEU &Success\\
\midrule
Original &0.681/0.569/0.620 &0.635/0.501/0.502 &3.79 &0.268 \\
Cleaned  &0.686/0.595/0.637 &0.656/0.547/0.596 &4.13 &0.279\\
\bottomrule
\end{tabular}
}
% \vspace{-0.8em}
\caption{The results of Task 2 baseline on the official dev set. U-P/R/F1 and S-P/R/F1 denote P/R/F1 for the user side and the system side, respectively.}
\label{tab:task2-result}
\end{table}
It can be seen that the model trained on the cleaned dataset outperforms the model trained on the original dataset in all metrics, which demonstrates the benefit of cleaning up redundant conversations.
Nevertheless, the results on the cleaned MobileCS still fall behind by a large margin in comparison to the results on other Wizard-of-Oz datasets.
For example, the Success rate of state-of-the-art models on MultiWOZ2.1 is around 75\%, while it is lower than 30\% on MobileCS.
The BLEU score on MobileCS is much lower than that on CrossWOZ \cite{liu2021variational}. Note that both TOD systems on MobileCS and CrossWOZ are based on Chinese GPT-2, though not strictly comparable.
% Some metrics of both models (especially BLEU score) are very low.
These results demonstrate how challenging of building TOD systems for real-life tasks is. The agent responses from real-life are much more difficult to be modeled, as compared those in the Wizard-of-Oz scenarios.

We further perform human evaluation for the best baseline model (i.e. the model trained on the cleaned dataset) and the average scores of all tested dialogues are shown in Table~\ref{tab:task2-human}. The scores of the three metrics are relatively low (lower than 3), which shows that in most cases, responses generated by the baseline system are neither fluent nor coherent enough, and can not provide requested information satisfactorily. In a word, building a TOD system that can perform well on real-life dialogues
is very challenging, and there is much room for the baseline TOD system to be improved. The MobileCS dataset offers a valuable and challenging testbed for future research of building human-robot dialogue systems for real-life tasks.
\begin{table}[t]
\centering
\resizebox{0.7\linewidth}{!}{
\begin{tabular}{ccc}
\toprule
Fluency & Coherency & Success\\
\midrule
2.76 &2.18 &2.24 \\
\bottomrule
\end{tabular}
}
% \vspace{-0.8em}
\caption{Human evaluation of the Task 2 baseline system (trained on the cleaned dataset).}
\label{tab:task2-human}
\end{table}

\section{Discussion and Conclusion}
% This paper describes the characteristics of MobileCS, a dataset from real life, which is distinct from previous TOD datasets. To promote studies on this real-life dataset, two tasks, information extraction and task-oriented dialog system, are proposed. However, dealing with these dialogues from real life and building baseline systems are extremely challenging. This paper then introduces the problems we meet and the solutions we propose when building baseline baseline systems.
% From the results of both tasks, we can see that to extract dialogue information and build dialogue systems on a real-life datasets are extremely challenging and need further investigation.
The performance of task-oriented dialogue systems on Wizard-of-Oz datasets have been improved continuously to a high level, for example, as shown in MultiWOZ\footnote{https://github.com/budzianowski/multiwoz}.
However, Wizard-of-Oz dialogue data are in fact simulated data and thus are fundamentally different from real-life conversations, which are more noisy and casual.
For further advancement of human-robot dialogue technology, real human-human dialogue data with grounding annotations to KBs are highly desirable.
Further, noting that the KB is an indispensable part for TOD systems and usually is not readily available for real-life tasks, it is very important to investigate not only the dialogue system itself but also information extraction to construct the KB.

With the MobileCS dataset released by the SereTOD challenge, this paper presents a baseline study of both information extraction (Task 1) and human-robot dialogue (Task 2) over real human-human dialogue data. We introduce how the baselines for the two tasks are constructed, the problems encountered, and the results.
It is found that the MobileCS dataset offers a challenging testbed for both tasks, with interesting open problems. Our baselines provide an easy entry to investigate the new dataset, and we anticipate that the baselines can facilitate exciting future research to build human-robot dialogue systems for real-life tasks.

% Entries for the entire Anthology, followed by custom entries
\bibliography{anthology,custom, addition_bib}
\bibliographystyle{acl_natbib}

\appendix

% \section{Example Appendix}
% \label{sec:appendix}

% This is an appendix.

\end{document}